\documentclass[10pt,twocolumn,letterpaper]{article}

\usepackage[pagenumbers]{cvpr}

\definecolor{cvprblue}{rgb}{0.21,0.49,0.74}
\usepackage[pagebackref,breaklinks,colorlinks,allcolors=cvprblue]{hyperref}

\usepackage{booktabs}
\usepackage{multirow}
\usepackage{graphicx}
\usepackage{pifont}
\usepackage{marvosym}
\newcommand{\cmark}{\ding{51}}
\newcommand{\xmark}{\ding{55}}
\newcommand{\pmark}{$\triangle$}
\newcommand{\method}{Real2Worlds2Real}

\title{
Real2Worlds2Real: Scaling Self-Improving Agentic Real2Sim2Real\\ for Robot Manipulation
\vspace{-3mm}
}

\AtBeginDocument{%
  \author{%
    Jinzhong Gu\textsuperscript{1,*}\enspace
    Jiakai Chen\textsuperscript{3,*}\enspace
    Xinyi Cao\textsuperscript{4,*}\enspace
    Yan Shen\textsuperscript{1,*}\enspace
    Hangke Sui\textsuperscript{3,*}\enspace
    Shutian Zheng\textsuperscript{1} \\
    Xiang Xiao\textsuperscript{2,\textdagger}\quad
    Haoran Lu\textsuperscript{1}\quad
    Fan Xu\textsuperscript{1}\quad
    Lu Shi\textsuperscript{2}\quad
    Daochen Shi\textsuperscript{2}\\
    Ruihua Zhang\textsuperscript{2}\quad
    Wei Zhan\textsuperscript{1}\quad
    Masayoshi Tomizuka\textsuperscript{1}\quad
    Ruihai Wu\textsuperscript{1,\Letter}\\[4pt]
    {\textsuperscript{1}University of California, Berkeley\quad
    \textsuperscript{2}NVIDIA}\\
    {\textsuperscript{3}University of Illinois Urbana-Champaign\quad
    \textsuperscript{4}Carnegie Mellon University}\\[2pt]
    {\small
      \textsuperscript{*}Equal contribution.\quad
      \textsuperscript{\textdagger}Lead author at NVIDIA.\quad
      \textsuperscript{\Letter}Corresponding author.}%
  }%
  \let\PublicOriginalMaketitle\maketitle
  \renewcommand{\maketitle}{%
    \begingroup
    \toggletrue{cvprfinal}%
    \PublicOriginalMaketitle
    \endgroup
  }%
}

\newcommand{\paperoverview}{%
\vspace{-7.6mm}
    \centering
    \includegraphics[width=1.0\textwidth]{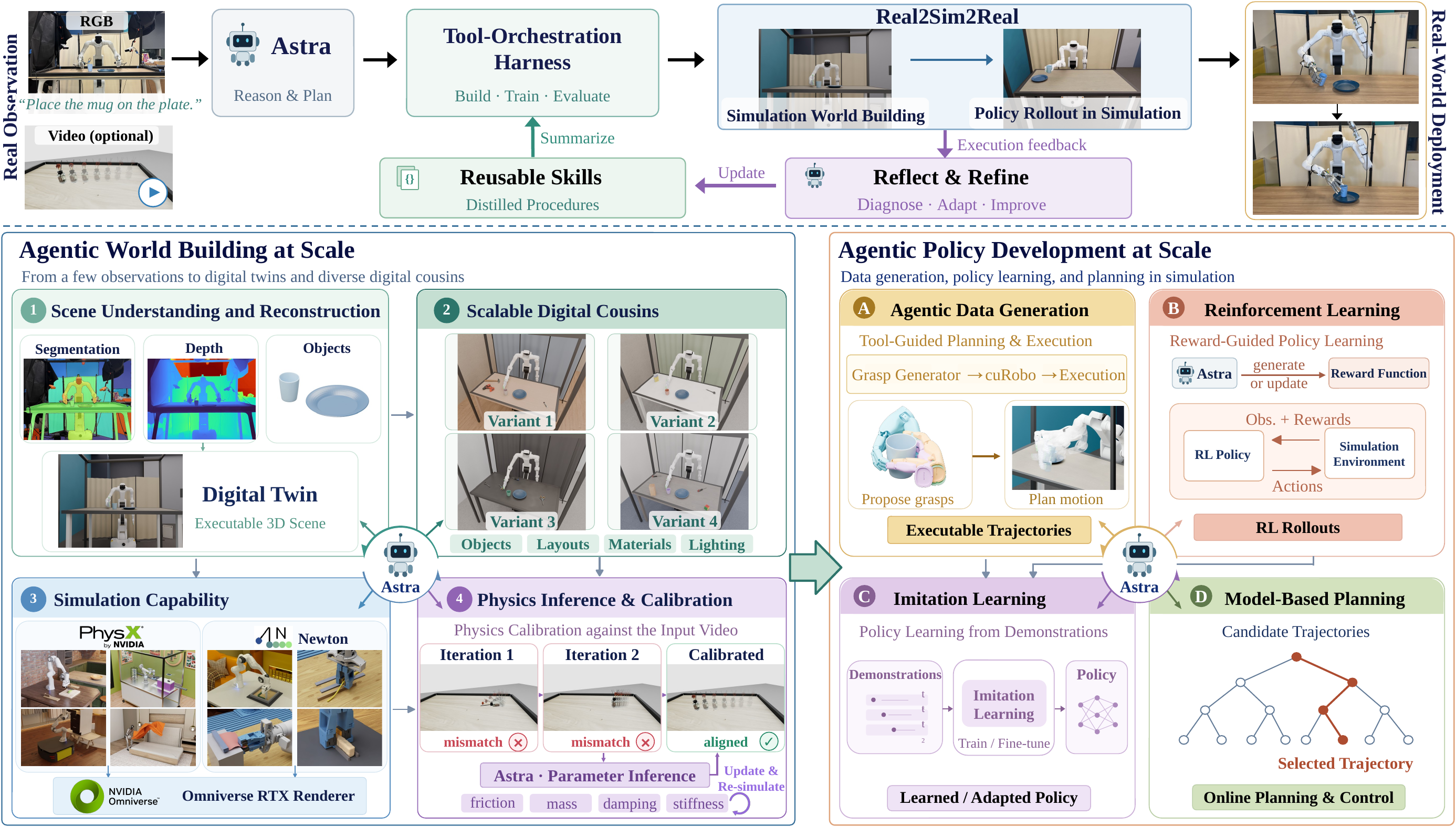}
    \par\vspace{1pt}
    \captionsetup{skip=0pt}
    \captionof{figure}{
    \textbf{Overview of Real2Worlds2Real.}
    An Astra-powered agentic framework connects world construction and policy development in an end-to-end Real2Sim2Real pipeline. It iteratively constructs and refines faithful digital twins and diverse, physically executable digital cousins, enabling scalable data generation and policy learning in simulation. The resulting policies are then deployed on real robots.
    }
    \label{fig:teaser}
    \vspace{4.3mm}
}

\makeatletter
\patchcmd{\@maketitle}{\vskip .375in}{\vskip .10in}{}{}
\patchcmd{\@maketitle}{\vspace*{24pt}}{\vspace*{13pt}}{}{}
\patchcmd{\@maketitle}{\vspace*{12pt}}{\vspace*{4pt}}{}{}
\apptocmd{\@maketitle}{%
  \par\noindent
  \begin{minipage}{\textwidth}
    \paperoverview
  \end{minipage}\par
}{}{\PackageError{paper}{Could not attach public overview to title}{}}
\makeatother

\begin{document}
\maketitle
\makeatletter
\ifdefined\makePagewiseLineNumber
  \renewcommand{\makePagewiseLineNumber}{%
    \logtheLineNumber\getLineNumber
    \if@firstcolumn\makeLineNumberLeft\else\makeLineNumberRight\fi
  }
  \let\makeLineNumber\makePagewiseLineNumber
\fi
\makeatother
\begingroup
\raggedbottom
\hyphenpenalty=300
\patchcmd{\abstract}{\vspace*{12pt}}{\vspace*{\baselineskip}}{}{}
\patchcmd{\endabstract}{\vspace*{12pt}}{\vspace*{0pt}}{}{}
\begin{abstract}
Scaling robot manipulation learning remains challenging due to the high cost of collecting real-world interaction data. Real-to-Sim-to-Real (Real2Sim2Real) offers a promising route by enabling policy learning in simulation grounded in real-world observations, yet scaling this paradigm across diverse tasks and physical interactions remains difficult. We introduce \textbf{Real2Worlds2Real}, a self-improving agentic framework that integrates scalable simulation-world construction with policy development. Given limited visual observations and task descriptions, its tool-orchestration harness constructs faithful digital twins and expands them into diverse, physically executable digital cousins. The agent adaptively selects tools and revisits construction decisions, using visual and execution feedback to refine geometry and physical behavior. Building on these worlds, the agent orchestrates policy development by coordinating data generation, policy learning, and refinement based on task requirements and simulation rollout feedback. Across both stages, the agent distills validated procedures into reusable skills that guide subsequent tasks. Experiments demonstrate improved world fidelity and successful real-world policy transfer, with performance gains from increasing world diversity and interaction-data volume. Skill reuse further improves world-building efficiency and long-horizon task success. Project website: \href{https://real2worlds2real.github.io/}{real2worlds2real.github.io}.
\end{abstract}

\clearpage
\endgroup
\section{Introduction}
\label{sec:intro}

Recent advances in robotic foundation models, including vision-language-action (VLA) policies~\cite{pmlr-v270-kim25c,BlackK-RSS-25,gr00tn1_2025} and world-action models (WAM)~\cite{cen2025worldvla,kim2025cosmospolicy,ye2026worldactionmodelszeroshot}, have demonstrated the importance of large and diverse data for generalizable robot manipulation. Yet collecting such data on physical robots remains costly, requiring substantial human effort for teleoperation, supervision, and repeated scene setup and resets.
These costs make it difficult to cover the breadth of manipulation tasks needed for generalization.
Real2Sim2Real offers a promising way to overcome this bottleneck. By grounding simulation in limited real-world observations, this paradigm offers the potential to scale both the diversity of training environments and the volume of robot interaction data, supporting policies that transfer to the real world.

Recent work has enabled the construction of visually realistic, interactive simulation environments from real observations~\cite{han2026re3sim,torne2024rialto,Jiang_2025_ICCV,jain2025polaris,jiang2026gsworld}, while automated scene generation and digital-cousin synthesis have expanded their diversity~\cite{dai2025digitalcousins,ranawaka2026simfoundry,chen2026agenticreal2sim,lu2026worldcomposer}. However, scaling Real2Sim2Real requires more than generating additional scenes.
Simulation worlds must faithfully reproduce task-relevant appearance and physical interactions while supporting diverse, physically meaningful variations that remain executable. Beyond world construction, effective robot learning also requires adapting data generation and policy refinement to different tasks, using failures observed in simulated rollouts to guide subsequent policy improvements. Together, these challenges motivate adaptive, feedback-driven approaches to both world building and policy development.

We introduce \textbf{Real2Worlds2Real}, a self-improving agentic Real2Sim2Real framework that transforms limited real-world observations into faithful digital twins and diverse, physically executable digital cousins for scalable robot policy development (Fig.~\ref{fig:teaser}).
For simulation-world construction, a \emph{tool-orchestration harness} enables a general-purpose agent to adaptively coordinate world building across diverse scenes and physical interactions. Guided by visual and execution feedback, the agent iteratively identifies and corrects errors in scene geometry and physical behavior. Successful correction procedures can be distilled into reusable skills, allowing experience from previous scenes to guide subsequent world building.
Starting from RGB images or videos, the agent reconstructs a scene-level digital twin that faithfully captures the real-world appearance and spatial structure.
It then creates a diverse family of digital cousins by introducing variations at both the object and scene levels, while ensuring that the resulting environments remain plausible and suitable for the intended manipulation tasks.
Beyond rigid objects, our framework supports the generation and physics-based simulation of articulated and deformable objects, enabling realistic robot interactions across a broader range of manipulation scenarios.
To further improve physical fidelity, when interaction recordings are available, the harness compares simulated rollouts with observed real-world motion, using their discrepancies to calibrate physical parameters.
These capabilities enable high-fidelity simulation world building to scale across diverse scenes and physical interactions without scene-specific human intervention.

Building on the simulation worlds, Real2Worlds2Real employs agentic tool orchestration for robot policy development. Rather than following a fixed training pipeline, the agent adaptively selects tools and learning strategies based on task requirements and available demonstrations.
It can synthesize executable control programs through tool-assisted grasp and motion planning, train reinforcement learning policies with agent-designed rewards, or employ model predictive control for online planning.
These methods can directly execute tasks or generate demonstrations across diverse simulation environments for training visuomotor policies through imitation learning.
Through simulated rollouts, the agent evaluates policy performance and uses failure feedback to guide further data generation and policy refinement, iteratively improving the resulting policies.
Across iterations, validated development procedures are distilled into reusable skills, allowing accumulated experience to guide policy development on subsequent tasks.

We evaluate Real2Worlds2Real across diverse manipulation scenes and physical interactions, assessing both the quality of simulation worlds and their effectiveness for robot policy learning.
Real-robot experiments demonstrate that policies trained in simulation successfully transfer to physical robots.
World-quality evaluations show that constructed worlds faithfully reproduce real-world appearance and physical interactions.
We further identify two complementary scaling trends: increasing interaction data improves policy performance, while increasing world diversity enhances policy generalization under a fixed interaction-data budget.
Finally, we demonstrate self-improvement in both world building and policy development, showing that reusing skills from previous tasks reduces world-building token consumption and improves long-horizon task success.

To summarize, our contributions are threefold:
\begin{itemize}
\item We introduce \textbf{Real2Worlds2Real}, a scalable Real-to-Sim-to-Real framework that constructs faithful digital twins and diverse digital cousins from limited real-world observations for agent-orchestrated policy development.
\item We develop \emph{agentic tool-orchestration workflows} for both world building and policy development, enabling iterative refinement through feedback and cross-task self-improvement through skill reuse.
\item We demonstrate the transfer of simulation-developed policies to real robots, validate the visual and physical fidelity of constructed worlds, and reveal scaling trends in world diversity and interaction-data volume.

\end{itemize}

\section{Related Work}
\vspace{-1mm}

\subsection{Real-to-Sim Scene Generation}
\label{sec:rw_real2sim}
\vspace{-1mm}

\textbf{Reconstruction-based digital twins.}
Mesh- and Gaussian-based methods reconstruct photorealistic digital twins from real captures~\cite{torne2024rialto,qureshi2025splatsim,han2026re3sim,jiang2026gsworld,jain2025polaris}, but often require markers for metric alignment~\cite{han2026re3sim,jiang2026gsworld,jain2025polaris} or manual segmentation, articulation, and assembly~\cite{torne2024rialto,jain2025polaris}, with default or hand-tuned physics.
PhysTwin~\cite{Jiang_2025_ICCV} and subsequent work~\cite{zhang2025real2simeval} identify deformable dynamics from interaction videos, but require calibrated multi-view RGB-D capture and operate per object.
These approaches primarily reconstruct individual twins rather than diverse world distributions.

\textbf{Generative expansion and digital cousins.}
Digital-cousin methods scale scene generation by retrieving or generating assets that preserve observed semantics and layouts while varying geometry and appearance~\cite{dai2025digitalcousins,ranawaka2026simfoundry,lu2026worldcomposer}.
Other methods infer articulation~\cite{chen2024urdformer,le2025articulateanything}, lift images to 3D~\cite{sam3d2025}, or generate scenes from text~\cite{yang2024holodeck,wang2024robogen}.
Typically organized as fixed pipelines with geometric checks or physics settling, these systems offer limited mechanisms for revisiting upstream errors in scale or support geometry, while assigning physical parameters through defaults or VLM priors.

\textbf{Agentic and self-improving construction.}
Agentic Real2Sim~\cite{chen2026agenticreal2sim} constructs episodic twins from recordings using stage critics and grasp refinement; ARSTAG~\cite{li2026arstag} coordinates agents from one image and resamples layouts after failed rollouts.
Both retain fixed stage graphs and toolsets without identifying rigid-body physics from real observations.
Built on agents that accumulate skills~\cite{wang2023voyager}, refine rewards~\cite{ma2024eureka}, or improve code~\cite{zelikman2024stop,zhang2025dgm,novikov2025alphaevolve}, \method{} dynamically orchestrates tools, verifies worlds through task execution and real-observation replay, and retains newly authored, validated skills across scenes.

\vspace{-1mm}
\subsection{Real2Sim2Real Policy Learning}
\vspace{-1mm}
Simulation scales training beyond costly real-world demonstrations.
MimicGen and related methods augment demonstrations via object-centric adaptation and skill recombination~\cite{mandlekar2023mimicgen,garrett2025skillmimicgen,jiang2025dexmimicgen,wang2024cyberdemo,lin2025cpgen}.
RoboGen~\cite{wang2024robogen} and GenSim2~\cite{hua2025gensim2} generate tasks and supervision through planning and RL, while agentic frameworks automate trajectory discovery through hybrid solvers and skill exploration~\cite{gong2025anytask,zhu2026skillweaver}.
ARSTAG~\cite{li2026arstag}, DREAM~\cite{sato2026dream}, and DexAgent~\cite{wang2026dexagent} reconstruct task-specific simulations to generate demonstrations using state-based experts, task-and-motion planning, or human-motion guidance.
\method{} instead couples diverse, physically grounded worlds with agent-orchestrated data generation,
studying how world diversity and data scale affect policy performance and generalization.

\suppressfloats[t]
\begin{figure}[t]
    \centering
    \includegraphics[width=\columnwidth]{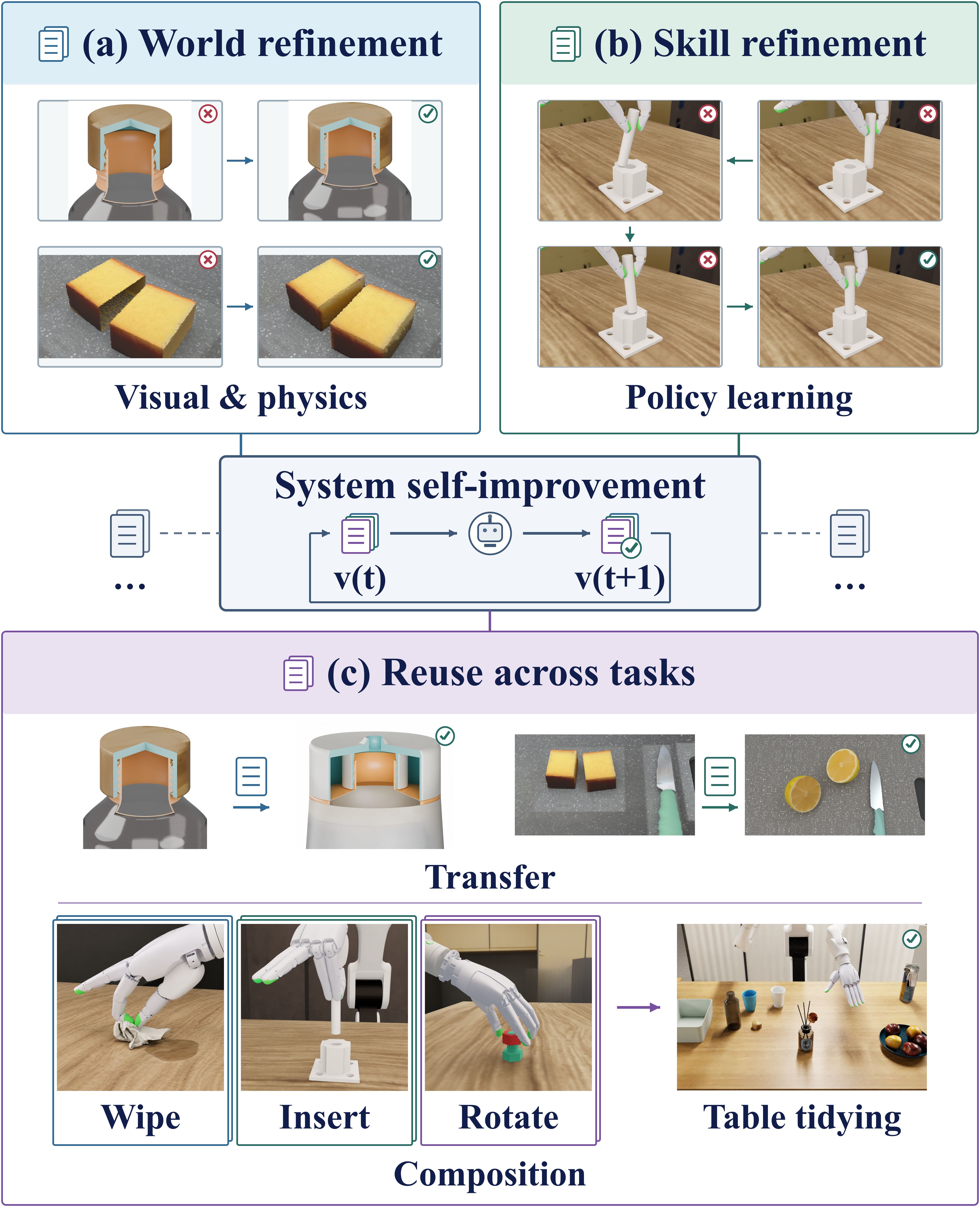}
    \vspace{-7mm}
    \caption{\textbf{Self-improvement through evolving reusable skills.} Both world-building and policy skills improve through feedback. Their reuse and composition enable new skills and tasks, further driving continual system self-improvement.}
    \vspace{-3mm}
    \label{fig:self_improvement}
\end{figure}

\begin{figure*}[htbp]
    \centering
    \includegraphics[width=1\linewidth]{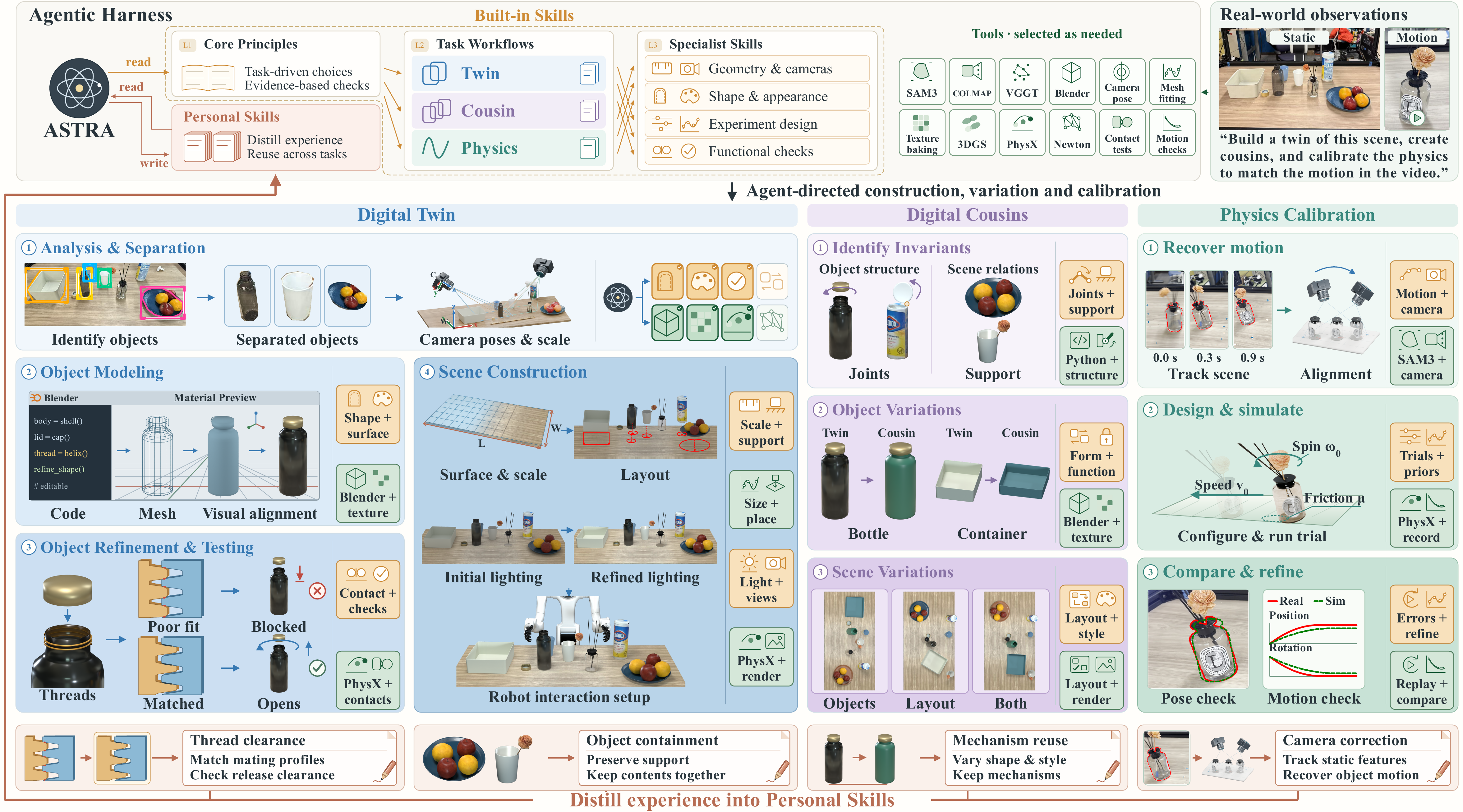}
    \vspace{-7mm}
    \caption{\textbf{Agentic simulation world building in Real2Worlds2Real.} Guided by built-in skills, the agent adaptively selects tools to reconstruct digital twins, generate task-consistent digital cousins, and infer and refine physical parameters. Visual and execution feedback guide iterative refinement, while effective world-building strategies are distilled into personal skills for reuse across tasks.}
    \label{fig:framework}
    \vspace{-3mm}
\end{figure*}

\section{Method}
Real2Worlds2Real is an agentic, self-improving  Real-to-Sim-to-Real framework that integrates scalable simulation-world construction with robot policy development.
As illustrated in Figure~\ref{fig:self_improvement}, execution feedback guides refinement across world construction and policy development, while effective procedures are distilled into reusable skills for future tasks.
Given RGB images or videos and a language description of a manipulation task, the agent first uses a skill-guided tool-orchestration harness to adaptively construct and refine simulation worlds (Sec.~\ref{subsec:harness}).
As shown in Figure~\ref{fig:framework}, it reconstructs a faithful digital twin and expands it into diverse, physically executable digital cousins, iteratively validating and refining their geometry and physical behavior (Sec.~\ref{subsec:world_building}).
Building on these worlds, the agent orchestrates policy development and refinement in simulation for deployment on real robots (Sec.~\ref{subsec:data_generation}).

\subsection{Agentic World-Building Self-Improvement}\label{subsec:harness}
We describe how the harness guides world-building decisions, supports iterative verification and refinement through visual and execution feedback, and preserves experience gained through refinement for reuse across future tasks.

\vspace{-4mm}
\paragraph{Skill-Guided Agentic Orchestration.}
The harness provides a general-purpose coding agent with a shared workspace that maintains the evolving scene and results from previous operations. This allows the agent to adapt its world-building decisions as construction progresses.
To guide this process, the harness organizes reusable procedural knowledge into three levels.
\emph{Core Principles} guide the agent to prioritize task requirements, distinguish observed properties from inferred ones, and verify generated results against the available evidence. \emph{Task Workflows} define high-level procedures, constraints, and acceptance requirements for world-building stages such as digital-twin reconstruction or digital-cousin generation. \emph{Specialized Skills} provide focused guidance for particular operations, including geometric reconstruction, articulation modeling, and physical simulation.
At task initialization, the agent receives the core principles and the relevant task workflow, and consults specialized skills as needed based on task requirements and execution feedback.
Together, these levels guide the agent's decisions without prescribing a fixed sequence of tool calls.

\vspace{-4mm}
\paragraph{Execution-Grounded Refinement and Verification.}
During world building, the agent iteratively refines candidate worlds using visual and execution feedback. It compares rendered views with real-world observations to identify visual discrepancies, while simulated interactions reveal failures in physical behavior. The harness also guides these refinements through explicit task constraints. For example, an articulated digital twin must represent the necessary parts and joint relationships. The harness validates candidate worlds through task-specific simulation checks, such as joint motion, contact behavior, and deformation integrity, requiring evidence consistent with the current scene assets. When validation fails, the agent uses the diagnostics to revise relevant assets or physical parameters and reruns the checks.

\vspace{-4mm}
\paragraph{Experience Distillation and Skill Reuse.}
The experience gained during world building can also help the agent handle similar problems in future tasks. Guided by the core principles, the agent reviews execution records and distills effective construction and refinement procedures into \emph{personal skills}. These skills describe how a problem was identified and resolved, along with the evidence supporting the correction. They may also describe when a procedure is applicable and how it may be adapted to new scenes. Personal skills are maintained separately from built-in skills and can be saved and loaded through the coding client's native capabilities. In subsequent tasks, the agent can refer to relevant personal skills and adapt the procedures to the new scene.

\subsection{Scalable Simulation World Building} \label{subsec:world_building}
As shown in Figure~\ref{fig:framework}, Real2Worlds2Real uses the tool-orchestration harness to build diverse simulation worlds while preserving task-relevant geometry and physical interactions.
The agent reconstructs digital twins, expands them into families of digital cousins, and iteratively validates and refines them through rendering and simulation feedback.

\subsubsection{Scene Reconstruction}
Given visual observations and a manipulation task, the agent reconstructs manipulable objects as editable 3D assets. It writes and iteratively refines Blender scripts to model object geometry and task-relevant interaction structures, such as articulated parts, joints, and deformable surfaces. When additional geometric cues are available, the agent can further use object masks and depth estimates to constrain visible geometry, calibrated multiview observations to refine parameterized models through silhouette fitting, and known-size references to establish metric scale.

The surrounding environment is reconstructed separately using meshes, Gaussian splats~\cite{gaussians}, or their combination. For visually complex backgrounds with sufficient image coverage, Gaussian splats preserve detailed appearance without explicitly modeling every surface, while meshes provide collision geometry for contact surfaces, such as tables and obstacles. When the Gaussian background contains a manipulable object, the harness removes splats around the object's original location. The agent then checks rendered views for residual artifacts or missing background content, adjusting the removal region or reconstruction as needed.
The resulting editable scene provides the basis for digital-cousin generation and physics simulation.

\subsubsection{Digital Cousins}
Starting from the reconstructed digital twin, the agent generates object-level variations by modifying the geometry, appearance, and material properties of task objects, while keeping the original twin as a reference. These variations retain task-relevant interaction structures, including joint types, link connectivity, and control roles for articulated objects, as well as attachment relationships for deformable objects. Alternatively, when suitable assets are available, the agent can retrieve them from existing libraries~\cite{lehome,garmentlab}.

At the scene level, the agent further expands diversity by combining object-level cousins with variations in scene layout, background appearance, and lighting.
For layout variations, the agent identifies important spatial and functional relationships from the reconstructed twin and manipulation task. It then repositions and reorients objects while maintaining stable support, sufficient space for robot manipulation, and task-relevant functional relationships.
Each candidate scene is evaluated through rendering and task-relevant simulated interactions to verify visual plausibility, physical stability, and manipulation feasibility before acceptance.

\subsubsection{Physics Simulation and Rendering}
The reconstructed assets are converted into physical models according to the interactions they must support. Rigid and articulated objects are simulated with PhysX~\cite{physxsdk}, using collision geometry and joint constraints to model contact and motion. For deformable objects, the harness supports physical representations suited to their behavior, including particle-based surfaces for cloth, rod-based models for ropes, and volumetric meshes for soft bodies. These representations are simulated using compatible PhysX or Newton~\cite{newtonphysics} solvers. Detailed visual surfaces are driven by the simulated deformation, enabling high-fidelity rendering without increasing the physical simulation resolution.
For rendering, the harness combines interactive mesh assets and their surrounding environment within a shared USD scene. Using Omniverse RTX Renderer~\cite{omniversertx}, it supports both conventional mesh-based rendering and hybrid rendering that combines mesh objects with Gaussian-splat backgrounds.

\subsubsection{Physics Inference and Calibration}
Beyond geometry and interaction structure, each simulation world requires physical parameters that govern object motion and contact, including mass, friction, damping, and material stiffness. The harness initializes these parameters from geometry and material estimates, for example, an object's mass can be estimated from its volume and assumed density. The agent can further evaluate the initialized models through task-relevant simulation, checking their physical plausibility and ability to support intended interactions, such as stable contact, valid joint motion, and reasonable deformation. When simulations reveal unexpected behavior, the agent revises the relevant physical parameters or model configurations and repeats the affected checks.

When real-world interaction recordings are available, the harness further calibrates physical parameters against observed motion in simulation.
The agent identifies parameters that can be constrained by the observed motion and designs corresponding simulation trials, specifying initial conditions, controls, and candidate parameter ranges to match the observed interactions. For each candidate, the harness resets the simulation, applies and verifies the effective parameter values, and records the resulting trajectories. These trajectories are then aligned with real-world observations to measure motion discrepancies, which guide subsequent parameter refinement.

\subsection{Agent-Orchestrated Policy Development}\label{subsec:data_generation}

Rather than following a fixed policy training pipeline, the agent adaptively orchestrates policy development based on task requirements and the demonstrations or policies already available.
It selects suitable tools and learning methods to construct policies or refine existing ones.
Throughout this process, feedback from simulated rollouts guides policy refinement and helps the agent determine whether additional demonstrations or further training are needed.

The agent can synthesize executable control programs in a \emph{code-as-policy} paradigm~\cite{lu2026aspire}.
These programs can segment and localize task objects from visual observations, and then coordinate manipulation by combining BODex-style grasp optimization~\cite{chen2025bodex} with motion planning through cuRobo~\cite{curobo_report23}, or invoking model predictive control (MPC)~\cite{mayne2000mpc} to update actions using state feedback.
It can also train state-based reinforcement learning (RL) policies with task-specific rewards constructed from task objectives and simulation states.
Simulation feedback helps the agent revise program logic and tool parameters or refine RL reward designs.
Both control programs and state-based RL policies can serve directly as task policies or generate demonstrations for subsequent learning.

The agent can train visuomotor policies such as ACT~\cite{zhao2023act} and Diffusion Policy~\cite{chi2023diffusionpolicy} using available or simulation-generated demonstrations.
By collecting simulated demonstrations across diverse digital cousins with varied initial conditions, it broadens the range of task configurations represented in the training data.
The agent evaluates these policies in simulation and uses observed failures to decide whether and how to refine them.
When a compatible expert can provide corrective actions at states visited by the current policy, the agent can apply DAgger-style refinement~\cite{ross2011dagger}, adding these corrections to the demonstration set for further training.
It can also train an RL decorator~\cite{yuan2025policydecorator} through simulated interactions, learning residual action corrections from task rewards while keeping the base policy frozen.
Subsequent rollouts assess the updated policies and guide further development.

Across iterations, the agent distills validated development procedures into reusable skills, recording when they apply and which failures they address.
These skills guide tool use and strategy selection on subsequent tasks.
Finally, selected policies are evaluated on real robots.

\section{Experiments}
\vspace{-1mm}

\label{sec:experiment}
We organize our experiments around four key questions, moving from downstream robot learning to the underlying world-building process:
(1) Can policies developed in the constructed simulation worlds successfully transfer to physical robots (Sec.~\ref{sec:real_transfer})?
(2) How do policy performance and generalization scale with interaction-data volume and world diversity (Sec.~\ref{sec:scaling})?
(3) How faithfully do the constructed worlds reproduce real-world appearance, geometry, and physical interactions (Sec.~\ref{sec:exp_harness})?
(4) Can the harness improve world-building on novel tasks by reusing accumulated experience  (Sec.~\ref{sec:harness_ablation})?

\vspace{-1mm}

\subsection{Real-World Deployment}
\vspace{-1mm}

\label{sec:real_transfer}
We first evaluate real-world performance on precision-demanding manipulation tasks. We conduct two complementary comparisons in Table~\ref{tab:harness_rl_post_training}, focusing on (a) the effectiveness of reconstructed simulation worlds for policy refinement and (b) the effectiveness of agentic policy development in simulation. All experiments are conducted on a physical Dexmate robot equipped with Sharpa dexterous hands and Orbbec RGB-D cameras. We report real-robot task success rates over 30 trials per task for each method.

\vspace{-3mm}
\paragraph{Baselines.}
We use two baseline groups to evaluate simulation world construction and agentic policy development.

\emph{Simulation World Construction.}
To isolate the effect of simulation environments, we adopt a common policy refinement setup. For each task, we first train an ACT policy~\cite{zhao2023act} on $60$ real-world demonstrations, as the shared initialization and imitation-learning baseline. Based on this policy, all methods use the same RL-decorator~\cite{yuan2025policydecorator} refinement procedure with equal amounts of data.
We compare against (1) \emph{Real-World Training}, which refines the policy using additional real-world interactions, (2) \emph{PolaRiS}~\cite{jain2025polaris} and (3) \emph{SimFoundry}~\cite{ranawaka2026simfoundry}, which construct interactive simulation environments from real-world observations. For a controlled comparison, all Real2Sim methods, including ours, use the same input videos and construct only digital twins.

\emph{Agentic Policy Development.}
With all methods using our simulation environments, we compare our agentic policy-development approach against (1) \emph{ASPIRE}~\cite{lu2026aspire} and (2) \emph{ENPIRE}~\cite{xiao2026enpire}, two agentic frameworks for automated robot policy development. All methods can utilize the provided $60$ real-world demonstrations within their respective strategies. We also include ACT trained on these demonstrations as the imitation-learning baseline.

\begin{table}[t]
    \centering
    \caption{Real-world success rates on precision-demanding tasks.}
    \label{tab:harness_rl_post_training}
    \footnotesize
    \setlength{\tabcolsep}{2pt}
    \renewcommand{\arraystretch}{1.15}

    \textbf{(a) Comparison of Simulation-World Construction}\\[2pt]
\begin{tabular*}{\columnwidth}{@{}p{0.40\columnwidth}@{\extracolsep{\fill}}ccccc@{}}
    \toprule
    \textbf{Method}
        & \textbf{Peg}
        & \textbf{Cup}
        & \textbf{Gear}
        & \textbf{Wipe}
        & \textbf{Cap} \\
    \midrule
    ACT (Imitation Learning)
        & 66.7\% & 53.3\% & 43.3\% & 46.7\% & 53.3\% \\
    \midrule
    Real-World Training
        & 86.7\% & 96.7\% & 86.7\% & 60.0\% & 93.3\% \\
    PolaRiS
        & 83.3\% & 66.7\% & 60.0\% & 50.0\% & 63.3\% \\
    SimFoundry
        & 73.3\% & 83.3\% & 40.0\% & 46.7\% & 56.7\% \\
    \textbf{Real2Worlds2Real}
        & \textbf{100.0\%} & \textbf{100.0\%} & \textbf{93.3\%}
        & \textbf{70.0\%} & \textbf{93.3\%} \\
    \bottomrule
\end{tabular*}

    \vspace{8pt}

    \textbf{(b) Comparison of Agentic Policy Development}\\[2pt]
\begin{tabular*}{\columnwidth}{@{}p{0.40\columnwidth}@{\extracolsep{\fill}}cccc@{}}
    \toprule
    \textbf{Method}
        & \textbf{Lift}
        & \textbf{PickPlace}
        & \textbf{Nut}
        & \textbf{Cup} \\
    \midrule
    ACT (Imitation Learning)
        & 56.7\% & 50.0\% & 40.0\% & 53.3\% \\
    \midrule
    ENPIRE
        & 76.7\% & 83.3\% & 53.3\% & 80.0\% \\
    ASPIRE
        & 90.0\% & 86.7\% & 63.3\% & 80.0\% \\
    \textbf{Real2Worlds2Real}
        & \textbf{100.0\%} & \textbf{90.0\%}
        & \textbf{73.3\%} & \textbf{100.0\%} \\
    \bottomrule
\end{tabular*}
\end{table}

\vspace{-3mm}
\paragraph{Tasks.}
(1) \textit{Peg insertion (Peg):} insert a peg into a hole.
(2) \textit{Cup stacking (Cup):} place a blue cup onto a white cup.
(3) \textit{Gear insertion (Gear):} insert a gear onto a board.
(4) \textit{Tissue retrieval and wiping (Wipe):} open a container, extract a tissue, and wipe the table.
(5) \textit{Bottle-cap tightening (Cap):} pick up a loose cap, place it on the bottle, and screw it tight.
(6) \textit{Cup lifting (Lift):} lift a light-blue cup at least 10\,cm above its starting height.
(7) \textit{Cup pick-and-place (PickPlace):} pick up a blue cup and place it upright to the right of the white cup.
(8) \textit{Nut tightening (Nut):} tighten a nut on a fixed bolt by turning it five full rotations.

\vspace{-3mm}
\paragraph{Results.}
As shown in Table~\ref{tab:harness_rl_post_training}(a), Real2Worlds2Real substantially outperforms Real2Sim baselines and achieves comparable or better performance than Real-World Training on all tasks. The limited improvement of PolaRiS and SimFoundry over the ACT base policy underscores the importance of accurate task-critical geometry for policy refinement. For example, SimFoundry reconstructs the gears without teeth and with inaccurate shaft spacing, so corrections learned in simulation do not transfer to the real assembly, preventing effective refinement. Beyond reconstruction fidelity, simulation exposes the policy to a wider range of object placements under the same data budget, helping explain its gains over real-world refinement.

Table~\ref{tab:harness_rl_post_training}(b) shows Real2Worlds2Real outperforms agentic baselines, with notable gains on Nut and Cup. The Nut task illustrates our advantage: ASPIRE and ENPIRE rely on scripted turning motions sensitive to nut placement variations, whereas our framework augments scripted execution with a closed-loop policy to adjust actions based on execution feedback, improving robustness to such variations.

\vspace{-1.5mm}
\subsection{Scaling Interaction Data and World Diversity}
\label{sec:scaling}
\vspace{-1.5mm}

\begin{table*}[t]
\centering
\caption{\textbf{World quality on real scenes.}
Methods receive their prescribed inputs from the same captures.
Completeness-aware quality scores penalize missing objects and false
positives; surface F-score measures reference-visible surface accuracy
and completeness.
Bold marks the best results.
Metric definitions and details are provided in the supplementary material.
}
\vspace{-2mm}
\label{tab:world_quality}
\setlength{\tabcolsep}{3pt}
\renewcommand{\arraystretch}{1.15}
\newcommand{\thead}[1]{\begin{tabular}{@{}c@{}}#1\end{tabular}}
\resizebox{\textwidth}{!}{%
\begin{tabular}{l*{11}{c}}
\toprule
\multirow{2}{*}{\textbf{Method}} &
\multicolumn{3}{c}{\textbf{Visual Fidelity}} &
\multicolumn{4}{c}{\textbf{Geometric Fidelity}} &
\multicolumn{2}{c}{\textbf{Physical Fidelity}} &
\multicolumn{2}{c}{\textbf{Scene-level Fidelity}} \\
\cmidrule(lr){2-4}
\cmidrule(lr){5-8}
\cmidrule(lr){9-10}
\cmidrule(lr){11-12}
& \thead{PSNR $\uparrow$}
& \thead{SSIM $\uparrow$}
& \thead{LPIPS $\downarrow$}
& \thead{Depth\\quality $\uparrow$}
& \thead{Object F1 $\uparrow$}
& \thead{Position\\quality $\uparrow$}
& \thead{Visible-surface\\F-score@1\,cm $\uparrow$}
& \thead{Stability $\uparrow$}
& \thead{Penetration\\quality $\uparrow$}
& \thead{Support\\recall $\uparrow$}
& \thead{Surface occupancy\\IoU $\uparrow$} \\
\midrule
ACDC
& 9.5 & 0.08 & 1.12
& 0.000 & 0.05 & 0.000 & 0.000
& 0.00 & 0.00
& 0.00 & 0.04 \\
SimFoundry
& 10.4 & 0.60 & 0.66
& 0.005 & 0.26 & 0.006 & 0.002
& 0.17 & 0.00
& 0.01 & 0.05 \\
PolaRiS
& 21.5 & 0.72 & 0.34
& 0.033 & 0.58 & 0.033 & 0.037
& 0.28 & 0.09
& 0.16 & 0.07 \\
\midrule
Ours w/o agent (fixed pipeline)
& 12.3 & 0.61 & 0.46
& 0.001 & 0.54 & 0.001 & 0.002
& 0.10 & 0.41
& 0.47 & 0.01 \\
Ours w/o adaptive refinement
& 15.7 & 0.66 & 0.46
& 0.035 & 0.37 & 0.047 & 0.012
& 0.24 & 0.14
& 0.23 & 0.01 \\
\textbf{Ours}
& \textbf{23.8} & \textbf{0.85} & \textbf{0.25}
& \textbf{0.063} & \textbf{0.69}
& \textbf{0.062} & \textbf{0.108}
& \textbf{0.63} & \textbf{0.55}
& \textbf{0.58} & \textbf{0.16} \\
\bottomrule
\end{tabular}%
}
\vspace{-2mm}
\end{table*}

We investigate how Real2Worlds2Real's simulation worlds support policy scaling with interaction-data volume and world diversity.
To isolate these effects from agentic policy development, we adopt Diffusion Policy~\cite{chi2023diffusionpolicy} as a common policy learner under a fixed training setup. Using the representative Stack Cups task, we vary each dimension independently to assess policy performance and generalization.

\begin{figure}[t]
    \centering
    \includegraphics[width=\linewidth]{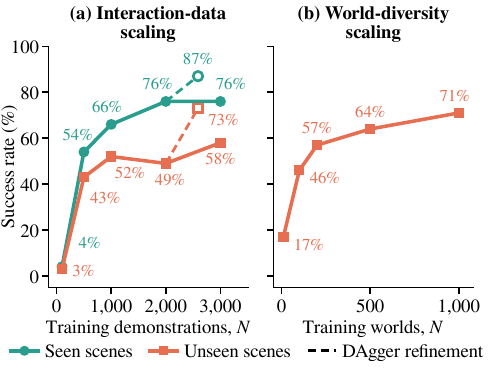}
    \vspace{-6mm}
    \caption{
    Scaling interaction-data volume and world diversity on the representative Stack Cups task.
    }
    \label{fig:scaling}
    \vspace{-2mm}
\end{figure}

\vspace{-3mm}
\paragraph{Interaction-Data Scaling.}
As shown in Fig.~\ref{fig:scaling}(a), we vary the number of training demonstrations across a fixed set of simulation worlds, with randomized object initial poses.
We evaluate the resulting policies on both seen and unseen scenes, observing that success rates initially improve but gradually plateau.
We further enable DAgger-based refinement, where the agent collects corrective demonstrations from failed rollouts and incorporates them into subsequent training.
This feedback-guided refinement yields additional performance gains beyond the plateau, highlighting the complementary benefits of increasing demonstration volume and learning from policy failures.

\vspace{-3mm}
\paragraph{World-Diversity Scaling.}
As shown in Fig.~\ref{fig:scaling}(b), we increase the number of training worlds from 10 to 1,000 via variations in object cousins, scene layouts, and backgrounds, while keeping the number of demonstrations fixed at 1,000. From the resulting policies on 100 unseen scenes, we observe consistent policy improvement as the number of training worlds increases, despite the fixed demonstration budget. The results reveal a clear scaling trend in policy generalization with increasing world diversity.

\subsection{World Fidelity and Agentic Design}
\label{sec:exp_harness}

\begin{table}[t]
    \centering
    \caption{\textbf{Physics parameter estimation from a single RGB video.}
    We evaluate parameter accuracy, motion prediction, and physical consistency; details are in the supplementary material.}
    \vspace{-2mm}
    \label{tab:physics_calibration}
    \footnotesize
    \setlength{\tabcolsep}{2pt}
    \renewcommand{\arraystretch}{1.1}
    \begin{tabular*}{\columnwidth}
    {@{}p{0.38\columnwidth}@{\extracolsep{\fill}}ccc@{}}
        \toprule
        \textbf{Method}
        & \shortstack{\textbf{Param.}\\\textbf{Err.} $\downarrow$}
        & \shortstack{\textbf{Predict.}\\\textbf{Err.} $\downarrow$}
        & \shortstack{\textbf{Bound}\\\textbf{Viol.} $\downarrow$} \\
        \midrule
        \multicolumn{4}{@{}l}{\textit{Default parameters}} \\
        PhysX default
        & 119.4\%
        & 83.1\%
        & 20.1\% \\
        LLM handbook values
        & 25.0\%
        & 18.3\%
        & 20.1\% \\
        \midrule
        \multicolumn{4}{@{}l}{\textit{Video-based system identification}} \\
        NIRPI~\cite{hofherr2023nirpi}
        & 48.4\%
        & 17.1\%
        & -- \\
        Garcia et al.~\cite{garcia2025learning}
        & 3702\%
        & 515\%
        & -- \\
        \midrule
        \multicolumn{4}{@{}l}{\textit{Single-pass inference}} \\
        2D tracking + analytic
        & 24.4\%
        & 13.1\%
        & -- \\
        Qwen3-VL-8B
        & 48.2\%
        & 23.6\%
        & 33.9\% \\
        Claude Opus 5.5
        & 25.4\%
        & 15.4\%
        & 20.0\% \\
        \midrule
        \textbf{Ours}
        & \textbf{4.2\%}
        & \textbf{2.6\%}
        & \textbf{0.0\%} \\
        \bottomrule
    \end{tabular*}
\end{table}

\paragraph{World Quality.}
Table~\ref{tab:world_quality} evaluates the visual, geometric, physical, and scene-level fidelity of reconstructed simulation worlds across real captures from Fruits~\cite{ranawaka2026simfoundry}, HOPE-Video~\cite{lin2021fusion,tyree2022hope}, and ARKitScenes~\cite{dehghan2021arkitscenes}. Each method receives its prescribed input and is scored without post-hoc alignment or rescaling.
Specifically, \emph{visual fidelity} measures reference-view rendering quality (PSNR, SSIM, and LPIPS); \emph{geometric fidelity} assesses depth accuracy, object detection completeness (Object F1), 3D placement accuracy (Position quality), and reference-visible surface reconstruction (Visible-surface F-score@1\,cm); \emph{physical fidelity} evaluates dynamic stability under gravity and collision-free placement in PyBullet (Stability and Penetration quality); and \emph{scene-level fidelity} measures inter-object support relationships and spatial layout consistency (Support recall and Surface occupancy IoU).
We further ablate the two core components: w/o agent replaces agentic orchestration with a fixed pipeline, and w/o adaptive refinement removes the feedback-driven refinement loop.
Across all four dimensions, \method{} consistently outperforms Real2Sim baselines and our ablated variants, confirming that both agentic orchestration and adaptive refinement are essential for generating faithful worlds.

\paragraph{Physics Calibration.}
We construct seven parameter-estimation tasks covering friction,
restitution, and pendulum length from real videos in
EV-RealPhys~\cite{kandukuri2024physics},
GAUGE~\cite{wang2026gauge}, and IRIS~\cite{khanbayov2026iris}.
Under a shared protocol, we compare default parameters,
video-based system identification, 2D tracking with closed-form
dynamics, and direct VLM inference. We assess agreement with
reference values and motion prediction in held-out trials.
As shown in Table~\ref{tab:physics_calibration}, our method covers
all seven tasks, achieving 4.2\% parameter error and 2.6\%
prediction error. Its friction bounds satisfy observed motion
constraints, and estimates vary little on the evaluated
perturbation subset. Task coverage varies across methods;
evaluation details are provided in the supplementary material.

\begin{figure}[htbp]
      \centering
      \begin{subfigure}[t]{\linewidth}
          \centering
          \includegraphics[width=\linewidth]{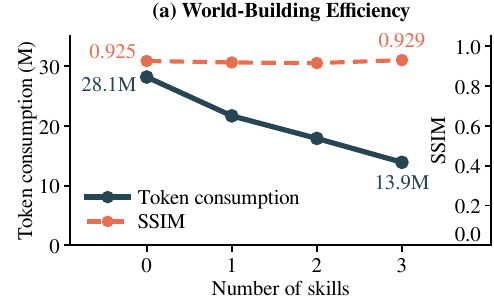}
          \label{fig:real2sim_SI}
      \end{subfigure}\par\vspace{1.5mm}
      \vspace{-3mm}
      \begin{subfigure}[t]{\linewidth}
          \centering
          \includegraphics[width=\linewidth]{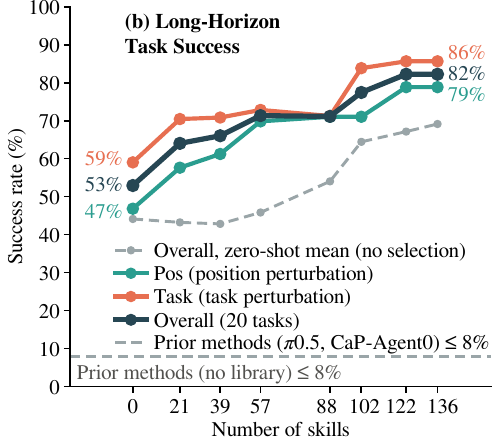}
          \label{fig:skill_scaling}
      \end{subfigure}
      \vspace{-5mm}
      \caption{
        \textbf{Benefits of scaling the skill library.}
        (a) Expanding the skill library reduces world-building token consumption from $28.1$M to $13.9$M while maintaining visual fidelity.
  (b)  LIBERO-Pro Long success increases from 52.9\% to 82.2\%, with improvements under both position and task perturbations.
  }
  \vspace{-5mm}
      \label{fig:skill_scaling_all}
\end{figure}

\subsection{Self-Improvement through Skill Reuse}
  \label{sec:harness_ablation}

  We test whether skills distilled from prior experience improve both ends of the Real2Sim2Real loop: building worlds and developing policies. In each experiment, we progressively expand the skill library while keeping the inputs, evaluation criteria, and other settings fixed.

  \paragraph{World-building efficiency.} We test whether experience from other tasks transfers to a new one, using lemon cutting as a held-out task. The agent builds the lemon-cutting world under four nested skill libraries: starting from the base skills, each extends the previous one with skills distilled from related tasks, namely fruit-handling experience, a geometric-cutting skill, and a full interactive-cutting skill. Inputs and verification criteria are identical across runs. As shown in Fig.~\ref{fig:skill_scaling_all}(a), the tokens needed to pass verification fall at every level, from 28.1M to 13.9M (about 50\%), while reconstruction quality remains comparable (SSIM of 0.925 vs.\ 0.929), indicating that the agent reuses stored procedures instead of rediscovering them by trial and error.

  \paragraph{Long-horizon task success.} On LIBERO-Pro Long~\cite{zhou2025liberopro}, the agent writes control scripts for 20 long-horizon tasks, half with position and half with task perturbations, each evaluated on 50 held-out initial states. As the library grows from 0 to 136 skills acquired on other tasks, overall success rises from 52.9\% to 82.2\% (Fig.~\ref{fig:skill_scaling_all}(b)), with gains under both position (46.8\% to 78.8\%) and task perturbations (59.0\% to 85.6\%). Even the mean over all candidate programs, without selecting the best, rises from 44.1\% to 69.1\%, showing that the skills yield better programs rather than better picks. In contrast, prior methods without a skill library ($\pi_{0.5}$~\cite{intelligence2025pi_}, CaP-Agent0~\cite{fu2026cap}) stay at or below 8\%.

\section{Conclusion}

We presented \method{}, a self-improving agentic framework for scaling Real2Sim2Real. From limited real-world observations, its harness constructs faithful digital twins and diverse, physically executable digital cousins, supporting adaptive data generation and policy development. Visual and execution feedback guide world and policy refinement, while successful procedures are distilled into reusable skills for future tasks. Experiments demonstrate improved world fidelity and effective policy transfer to real robots, alongside performance and generalization gains from increasing interaction data and world diversity. Accumulated skills further improve world-building efficiency and long-horizon task success, highlighting the value of experience reuse across the pipeline.

\section*{LLM Usage Statement}

We used GPT-6 Astra within the proposed tool-orchestration framework for simulation-world construction and simulation-driven policy development. Qwen3-VL-8B and Claude were used as comparison models in the video-based physical-parameter estimation experiments.

ChatGPT was used to assist with language polishing and LaTeX table formatting. All scientific claims, experimental results, and final manuscript content were verified by the authors, who take full responsibility for this work.

{
    \small
    \bibliographystyle{ieeenat_fullname}
    \bibliography{main}
}

\clearpage
\appendix
\setcounter{figure}{0}
\setcounter{table}{0}
\setcounter{equation}{0}
\renewcommand{\thefigure}{S\arabic{figure}}
\renewcommand{\thetable}{S\arabic{table}}
\renewcommand{\theequation}{S\arabic{equation}}
\renewcommand{\theHfigure}{S\arabic{figure}}
\renewcommand{\theHtable}{S\arabic{table}}
\renewcommand{\theHequation}{S\arabic{equation}}

\twocolumn[{
  \centering
  \Large
  {\renewcommand{\vspace}[1]{}\textbf{\thetitle}}\\
  \vspace{0.5em}Supplementary Material \\
  \vspace{1.0em}
  \begin{minipage}{\textwidth}
    \centering
    \captionsetup{hypcap=false}
    \includegraphics[width=\textwidth]{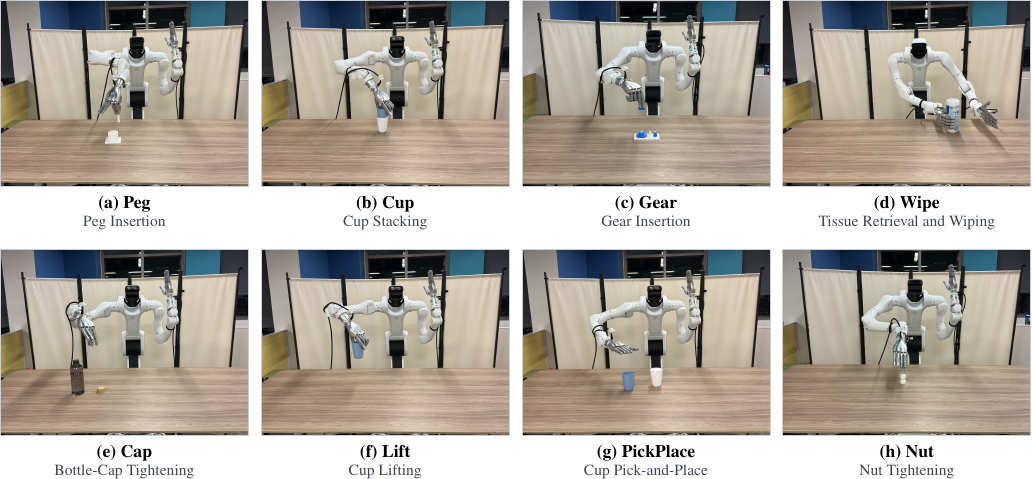}
    \captionof{figure}{\textbf{Real-world task gallery.} Overview of the eight precision-demanding real-robot manipulation tasks evaluated in Table~1(a) and Table~1(b) of the main paper.}
    \label{fig:real_task_gallery}
  \end{minipage}
  \vspace{1.2em}
}]

\section{Real-Robot Experiments and Task Gallery}
\label{sec:supp_real_exp}

Fig.~\ref{fig:real_task_gallery} illustrates the eight real-world manipulation tasks evaluated in Sec.~4.1 (Table~1(a) and Table~1(b)) of the main paper. These tasks encompass a range of contact-rich and precision-demanding manipulation skills, such as tight-tolerance insertion, multi-turn screwing, and bottle-cap alignment and fastening.

\section{Task Generalization with Digital Cousins}
\label{sec:supp_cousin}

We evaluate how well policies trained on generated data generalize on two simulated tasks: \emph{Stack Cups}, the representative rigid-object task of Sec.~4.2 of the main paper, and \emph{Fold Cloth}, a deformable-object task.
Table~\ref{tab:supp_cousin} summarizes the results.

\subsection{Setup}

\paragraph{Stack Cups.}
In our simulation, Stack Cups requires the robot to stack a cup upright onto a plate.
Real2Worlds2Real reconstructs the digital twin of the lab scene and expands it into digital cousins that vary the cups, scene layout, background, and lighting.
The agent's control programs (Sec.~3 of the main paper) generate 2,000 demonstrations in the twin scene and eight seen cousin scenes, with randomized object poses, distractors, and lighting, as in Fig.~4(a) of the main paper.
A trial succeeds if, within 30\,s, the cup rests upright on the plate and the hand has moved away.
We evaluate one rollout per case on three test sets of 100 cases each:
\begin{itemize}
\item \emph{Twin}: 100 new initial states in the twin scene as reconstructed, with the original lighting, no distractors, and no camera perturbation. None of these states appears in training or in the other test sets.
\item \emph{Seen}: 100 test scenes generated from the twin scene and the eight seen cousin scenes, with tabletop distractors, randomized lighting, and randomized camera pose.
\item \emph{Unseen}: the 100 unseen scenes of Fig.~4, generated in the same way from two held-out cousin scenes.
\end{itemize}
ACDC and SimFoundry generate data only in the twin scene, whereas our data and MimicGen's span all seen scenes.
\emph{Twin} therefore compares all methods in a scene that every method trains in, which separates the quality of the generated data from scene coverage.
\emph{Seen} measures performance within our training distribution, and \emph{Unseen} measures generalization to new scenes.

\paragraph{Fold Cloth.}
We use LeHome~\cite{lehome} with four garment categories: long pants, shorts, long-sleeve tops, and short-sleeve tops.
Training uses 40 seen garments.
\emph{Seen} has 200 test cases, five per seen garment, and \emph{Unseen} has 40 cases on eight held-out garments.
Each garment is scaled by a factor in $[0.85, 1.15]$.
Success follows the LeHome criterion, with its distance thresholds scaled by the same factor, under a limit of 600 steps.
Our policy is trained on demonstrations collected on the seen garments.

\paragraph{Policy learning.}
All methods share one training recipe per task: Diffusion Policy~\cite{chi2023diffusionpolicy} for Stack Cups and its point-cloud variant DP3~\cite{ze20243d} for Fold Cloth.
For every method, the checkpoint is selected on a separate validation set of 54 rollouts for Stack Cups and 80 for Fold Cloth.
On both tasks, our policy is further refined with the DAgger-based refinement of Sec.~4.2 of the main paper: the agent collects corrective demonstrations at states visited by the policy and adds them to training.
Table~\ref{tab:supp_cousin} reports the refined policies; the Stack Cups policy is the one shown in Fig.~4(a).

\paragraph{Baselines.}
All baselines use our robot, our policy-training recipe, and our test sets.
They are trained only on their generated data, because refinement requires expert actions at states visited by the learned policy, which their generators cannot provide.
Table~\ref{tab:supp_cousin} therefore compares the complete pipelines.

\emph{MimicGen}~\cite{mandlekar2023mimicgen} isolates demonstration generation from world building.
It splits each source demonstration into object-centric segments.
In a new scene, it re-expresses each segment relative to the new object poses, interpolates between segments, executes the result in simulation, and keeps only successful trials.
For Stack Cups, the segments are grasping the cup, expressed in the cup frame, and placing it, expressed in the plate frame.
MimicGen receives nine source demonstrations from our expert, one per seen scene, and generates data in the same scene configurations as our demonstrations.
It therefore shares our training scenes but produces its demonstrations differently.
Because MimicGen targets parallel-jaw grippers, we transform the wrist trajectory and keep the source hand-joint commands within each segment.
For Fold Cloth, MimicGen receives one LeHome demonstration per seen garment.
Garments have no rigid pose, so each segment is expressed relative to the semantic keypoint that the gripper grasps, using a planar transform of translation, yaw, and garment scale.

\emph{ACDC}~\cite{dai2025digitalcousins} and \emph{SimFoundry}~\cite{ranawaka2026simfoundry} are world-building baselines and receive the same input as our method.
Both build tabletop scenes: SimFoundry reconstructs a single support surface, and ACDC assembles a scene from retrieved assets.
Neither yields a room where our robot can operate and that matches the test scenes, so both use our reconstructed twin room.
Both also use our plate, because success is defined relative to the plate.
Both take robot and object initial states from our sampler in 716 twin-scene configurations, so all methods face the same task distribution.
Everything else comes from their own pipelines.

ACDC matches the objects detected in the input to assets retrieved from a library and composes digital cousins from the closest matches.
Its default library cannot be exported, so it retrieves a twin cup and five cup cousins from public asset libraries.
It adds no distractors, because it places only the detected objects.
It generates demonstrations with scripted skills that use privileged state.
Its original skills do not support our dexterous hand, so we build them on our task-agnostic hand primitives.

SimFoundry reconstructs the task objects, generates five cup cousins with generative and 3D-generation models, and adds distractors.
It then generates demonstrations MimicGen-style, with domain randomization at replay.
For reproducibility, it uses open-source generative models in place of its proprietary default.
Its distractors come from public libraries and exclude cups, plates, and bowls, so they cannot be confused with the task objects.
Replay randomizes materials and environment lighting, and camera perturbation is limited to the range of our test sets.
Following its own pipeline, it does not vary the room.
It uses the same nine source demonstrations as MimicGen.

ACDC and SimFoundry are each trained and evaluated with their own twin cup, with the checkpoint selected on validation with that cup. The number of training and validation samples used are the same as our method.
Neither supports deformable objects, so both are evaluated only on Stack Cups.

\begin{table}[t]
    \centering
    \caption{Simulation success rates on twin, seen, and unseen scenes.
    Stack Cups has 100 cases per test set.
    Fold Cloth has 200 \emph{Seen} cases and 40 \emph{Unseen} cases.
    ACDC and SimFoundry use their own twin cups; the other methods use ours.
    --: not supported.}
    \label{tab:supp_cousin}
    \footnotesize
    \setlength{\tabcolsep}{2pt}
    \renewcommand{\arraystretch}{1.15}
\begin{tabular*}{\columnwidth}{@{}p{0.28\columnwidth}@{\extracolsep{\fill}}ccccc@{}}
    \toprule
        & \multicolumn{3}{c}{\textbf{Stack Cups}}
        & \multicolumn{2}{c}{\textbf{Fold Cloth}} \\
    \cmidrule(lr){2-4}\cmidrule(l){5-6}
    \textbf{Method}
        & \textbf{Twin}
        & \textbf{Seen}
        & \textbf{Unseen}
        & \textbf{Seen}
        & \textbf{Unseen} \\
    \midrule
    MimicGen
        & 67.0\% & 59.0\% & 46.0\% & 65.0\% & 42.5\% \\
    ACDC
        & 10.0\% & 1.0\% & 0.0\% & -- & -- \\
    SimFoundry
        & 32.0\% & 11.0\% & 4.0\% & -- & -- \\
    \textbf{Real2Worlds2Real}
        & \textbf{93.0\%} & \textbf{87.0\%} & \textbf{73.0\%}
        & \textbf{83.0\%} & \textbf{55.0\%} \\
    \bottomrule
\end{tabular*}
\end{table}

\subsection{Results}

\paragraph{Generalization across scenes.}
Real2Worlds2Real outperforms every baseline on every test set of both tasks.
On Stack Cups, our policy goes from 93.0\% on \emph{Twin} to 87.0\% on \emph{Seen} and 73.0\% on \emph{Unseen}.
Over the same sets, SimFoundry drops from 32.0\% to 11.0\% and 4.0\%, and ACDC drops from 10.0\% to 1.0\% and 0.0\%.
Both baselines vary objects and appearance around a single twin scene, which does not prepare their policies for new layouts, backgrounds, and lighting.
Our scene-level cousins cover these variations, so our policy keeps most of its twin-scene performance in new scenes.
On Fold Cloth, our policy reaches 83.0\% on \emph{Seen} and 55.0\% on \emph{Unseen}, against 65.0\% and 42.5\% for MimicGen.

\paragraph{Data quality.}
The \emph{Twin} results show that scene coverage alone does not explain the gaps.
MimicGen generates data in the same training scenes as ours, yet it trails our policy by 26.0, 28.0, and 27.0 percentage points on \emph{Twin}, \emph{Seen}, and \emph{Unseen}.
Two properties of our data account for this gap.
First, our control programs plan each grasp and placement for the current object and scene, whereas MimicGen transforms a few source demonstrations and thus covers fewer grasp and placement strategies.
Even without refinement, our policy trained on 1,000 demonstrations (Fig.~4(a)) already reaches 66.0\% on \emph{Seen} and 52.0\% on \emph{Unseen}, above MimicGen's 59.0\% and 46.0\%.
Second, because the control programs plan from the current state, the agent can also supply expert actions at states visited by the learned policy, which enables DAgger-based refinement.
On Stack Cups, this raises our 2,000-demonstration policy from 76.0\% to 87.0\% on \emph{Seen} and from 49.0\% to 73.0\% on \emph{Unseen} (Fig.~4(a)).
The baselines lack this ability: MimicGen and SimFoundry build each demonstration from fixed source segments, and ACDC's scripted skills run from the scene's initial state, so none can correct the policy at the states it actually reaches.
ACDC and SimFoundry reach only 10.0\% and 32.0\% even on \emph{Twin}, the scene they train in, against 67.0\% for MimicGen.

\paragraph{Checks on the baseline settings.}
We verify that the low ACDC and SimFoundry results do not stem from our evaluation choices.
First, cup choice: evaluating them with our cups instead of their own twin cups gives 9.0\% (\emph{Seen}) and 3.0\% (\emph{Unseen}) for SimFoundry and 0.0\% on both for ACDC, so the cup is not the cause.
Second, distractors: ACDC trains without distractors, yet removing all distractors from the test scenes still gives 0.0\% on both sets.
Inspecting the rollouts, ACDC fails mostly during grasping, where the hand closes before contact or knocks the cup over.

\section{Extended Related Work and Comparisons}
\label{sec:supp_related_details}

\begin{table*}[htbp]
\centering
\caption{\textbf{Comparison of Real-to-Sim world-construction methods.}
\emph{Twin} denotes reconstruction of the observed scene.
\emph{Cousins} denotes variation in object geometry or appearance
beyond pose randomization. Triangles mark partial support, such as
a background-only or single-object twin.
\emph{Rig./Art./Def.} denote rigid, articulated, and deformable objects.
\emph{Calibrated} parameters are fitted to real observations.
\emph{Optim.} denotes prescribed per-object optimization.
\emph{Fixed} workflows use a predefined pipeline, with model queries
at designated steps where indicated. \emph{Staged agents} operate
within a fixed stage graph. \emph{Open-ended} agents select tools and
their order. Feedback codes are G (geometry or rule checks), V (visual
checks), S (simulated task execution), and R (alignment with real
observations). A dash means no feedback category is listed.
\emph{Self-improvement} denotes reuse of validated agent-authored procedures
across scenes. \emph{Human-free} denotes construction without
per-scene manual steps. Entries follow the original papers and our
reproductions.}
\vspace{-2mm}
\label{tab:related}
\footnotesize
\setlength{\tabcolsep}{2.4pt}
\begin{tabular}{@{}llccccclllcc@{}}
\toprule
 & & \multicolumn{5}{c}{World construction} & Physics & \multicolumn{4}{c}{Autonomy} \\
\cmidrule(lr){3-7}\cmidrule(lr){8-8}\cmidrule(l){9-12}
Method & Input & Twin & Cousins & Rig. & Art. & Def. & Source & Orchestration & Feedback & Self-Improvement & Human-free \\
\midrule
\multicolumn{12}{@{}l}{\textit{Reconstruction-based digital twins}} \\
RialTo~\cite{torne2024rialto}            & Video scan      & \cmark & \xmark & \cmark & \cmark & \xmark & Default    & Manual GUI  & --         & \xmark & \xmark \\
RE$^3$Sim~\cite{han2026re3sim}           & Images + markers& \cmark & \xmark & \cmark & \xmark & \xmark & Default    & Fixed       & --         & \xmark & \xmark \\
GSWorld~\cite{jiang2026gsworld}          & Video + markers & \pmark & \xmark & \cmark & \xmark & \xmark & Default    & Fixed       & --         & \xmark & \xmark \\
PolaRiS~\cite{jain2025polaris}           & Video + board   & \cmark & \xmark & \cmark & \xmark & \xmark & Manual     & Fixed + GUI & --         & \xmark & \xmark \\
PhysTwin~\cite{Jiang_2025_ICCV}        & RGB-D videos    & \pmark & \xmark & \xmark & \xmark & \cmark & Calibrated & Optim.      & R          & \xmark & \pmark \\
\midrule
\multicolumn{12}{@{}l}{\textit{Generative expansion and digital cousins}} \\
ACDC~\cite{dai2025digitalcousins}                  & Single RGB      & \xmark & \cmark & \cmark & \cmark & \xmark & Asset      & Fixed + VLM & G          & \xmark & \cmark \\
WorldComposer~\cite{lu2026worldcomposer} & Panorama        & \pmark & \cmark & \cmark & \pmark & \xmark & Asset      & Fixed + LLM & G          & \xmark & \xmark \\
SimFoundry~\cite{ranawaka2026simfoundry} & Video           & \cmark & \cmark & \cmark & \cmark & \xmark & VLM prior  & Fixed + VLM & G\,V       & \xmark & \cmark \\
\midrule
\multicolumn{12}{@{}l}{\textit{Agentic construction}} \\
Agentic Real2Sim~\cite{chen2026agenticreal2sim} & Robot rec. & \cmark & \pmark & \cmark & \xmark & \cmark & VLM prior & Staged agents & V\,S    & \xmark & \cmark \\
ARSTAG~\cite{li2026arstag}               & RGB + text      & \cmark & \pmark & \cmark & \xmark & \xmark & Default    & Staged agents & G\,V\,S & \xmark & \cmark \\
\midrule
\textbf{\method{}}                       & RGB / video     & \cmark & \cmark & \cmark & \cmark & \cmark & Calibrated & Open-ended  & G\,V\,S\,R & \cmark & \cmark \\
\bottomrule
\end{tabular}
\end{table*}

Real-to-Sim methods differ in what they recover from an observed scene,
how they expand it into training worlds, and what evidence they use to
revise those worlds. Table~\ref{tab:related} compares these choices for
representative construction workflows.

\paragraph{Digital twins and physical identification.}
Mesh- and Gaussian-based reconstruction provides a way to preserve the
appearance and spatial layout of a captured scene for robot
learning~\cite{torne2024rialto,qureshi2025splatsim,han2026re3sim,jiang2026gsworld,jain2025polaris}.
To make that scene interactive, the workflow must also establish metric
scale and represent the parts and contacts that govern manipulation.
RE$^3$Sim, GSWorld, and PolaRiS use markers or a calibration board for
metric alignment~\cite{han2026re3sim,jiang2026gsworld,jain2025polaris}.
RialTo and PolaRiS use manual steps for scene preparation or
assembly~\cite{torne2024rialto,jain2025polaris}. These choices describe
how a scene is built. Its physical parameters can still come from
simulator defaults or manual settings.

PhysTwin and related work take a system-identification approach to
deformable objects, fitting dynamics to observed interactions from
calibrated multi-view RGB-D
recordings~\cite{Jiang_2025_ICCV,zhang2025real2simeval}.
The scope of this fit explains the combination of a partial
\emph{Twin} entry and \emph{Calibrated} physics in the PhysTwin row.
A workflow can recover the dynamics of an individual object while
covering only part of the observed environment. This is why
Table~\ref{tab:related} separates scene coverage, supported object
types, and the source of physical parameters.

\paragraph{Digital cousins and task-consistent variation.}
Digital-cousin methods retrieve or generate assets that retain scene
semantics and layout while varying object geometry or
appearance~\cite{dai2025digitalcousins,ranawaka2026simfoundry,lu2026worldcomposer}.
ACDC illustrates asset-based variation from an observed scene, and
SimFoundry provides both scene reconstruction and cousin generation.
The table separates these two outputs because an asset can serve the
same role in a family of training scenes with varying degrees of
reconstruction of the original capture. The \emph{Cousins} column
credits changes to the objects themselves beyond pose randomization.
Variation confined to the background, lighting, or viewpoint is marked
as partial.

Related work addresses specific parts of construction, including
articulation recovery~\cite{chen2024urdformer,le2025articulateanything},
lifting individual images to 3D~\cite{sam3d2025}, and generating scenes
from text~\cite{yang2024holodeck,wang2024robogen}. For manipulation,
variation also has to preserve the relationships that make the task
possible, such as support, clearance, and joint motion. Geometric
checks and settling tests can detect some failures in a proposed scene.
Correcting their causes may require revisiting an earlier decision
about scale, object geometry, or support placement. In
\method{}, this is part of the refinement loop used to validate both
twins and cousins.

\paragraph{Agentic construction and skill reuse.}
Agentic Real2Sim builds episodic twins from robot recordings using
stage critics and simulation-based grasp
refinement~\cite{chen2026agenticreal2sim}. ARSTAG uses staged agents
and resamples layouts after failed rollouts~\cite{li2026arstag}.
These examples show that feedback can guide revision within a
predefined workflow. \emph{Orchestration} records who determines
which operation comes next, and \emph{Feedback} records the evidence
available to guide revision. \emph{Human-free} covers another
dimension. For example, SimFoundry automates scene construction
with model queries at predetermined points in the pipeline.

In \method{}, the workflow provides task requirements and verification
criteria, and the agent uses them to choose tools and revisit earlier
decisions. Rendered views and simulated task execution provide
complementary checks. When interaction recordings constrain the
physical parameters, discrepancies between observed and simulated
motion guide calibration. The \emph{Self-Improvement} column captures what
persists across scenes. Drawing on work on skill accumulation, reward
refinement, and program
improvement~\cite{wang2023voyager,ma2024eureka,zelikman2024stop,zhang2025dgm,novikov2025alphaevolve},
the harness distills validated construction procedures into reusable
skills. A later run can adapt a procedure to a similar problem using
the conditions and verification evidence recorded with that skill.

\paragraph{Connection to simulation-based policy learning.}
Demonstration augmentation methods expand supervision through
object-centric adaptation and skill
recombination~\cite{mandlekar2023mimicgen,garrett2025skillmimicgen,jiang2025dexmimicgen}.
Task-specific Real-to-Sim systems also generate demonstrations through
state-based experts, planning, or human-motion
guidance~\cite{li2026arstag,sato2026dream,wang2026dexagent}.
World construction determines which objects, layouts, and interactions
those demonstrations can cover. This motivates the separate scaling
experiments in the main paper, which vary interaction-data volume and
world diversity using a common policy learner.

\section{Visual and Physical Evaluation Metrics}
\label{sec:supp_metrics}

\subsection{World-Quality Evaluation}

\paragraph{Evaluation setting.}
We evaluate reconstructed worlds on real captures from
Fruits~\cite{ranawaka2026simfoundry},
HOPE-Video~\cite{lin2021fusion,tyree2022hope}, and
ARKitScenes~\cite{dehghan2021arkitscenes}.
Each method receives its prescribed input from the corresponding
capture. We compare RGB renders at the reference view and evaluate
exported geometry at its delivered pose and scale. Physical stability
and penetration are evaluated in a shared PyBullet setup.

\paragraph{Object matching and completeness.}
We match predicted objects one-to-one to reference instances using
visible silhouette IoU of at least $0.5$. Let $N_{\mathrm{GT}}$ denote
the number of reference objects and $N_{\mathrm{FP}}$ the number of
unmatched visible predictions. The depth, position, stability, and
penetration scores use the same scene-level normalization:
\begin{equation}
    Q = \frac{\sum_{i=1}^{N_{\mathrm{GT}}} q_i}
              {N_{\mathrm{GT}} + N_{\mathrm{FP}}}.
    \label{eq:supp_complete_quality}
\end{equation}
Missing or unverifiable objects receive $q_i=0$, and false positives
increase the denominator.

\paragraph{Visual fidelity.}
We compute PSNR, SSIM, and LPIPS over the full reference image after
resizing the render to the reference resolution. For RGB images
$I$ and $\widehat I$ with values in $[0,1]$,
\begin{equation}
    \mathrm{PSNR}
    = -10\log_{10}\!\left(\frac{\|I-\widehat I\|_2^2}{3HW}\right),
\end{equation}
where $H$ and $W$ are image height and width. SSIM measures local
agreement in luminance, contrast, and structure. LPIPS measures
perceptual distance using AlexNet features. These are reference-view
scores, and the reference view may be among the reconstruction inputs.

\paragraph{Geometric fidelity.}
For depth quality, let $\Omega_i$ be the valid reference-depth pixels
inside object $i$'s mask. Let $A_i\subseteq\Omega_i$ contain the pixels
where the frontmost rendered surface belongs to the matched object
and has valid depth. With reference depth $D$ and rendered depth
$\widehat D$, we compute
\begin{equation}
    q_i^{\mathrm{depth}}
    = \frac{1}{|\Omega_i|}
      \sum_{u\in A_i}
      \exp\!\left(-\frac{|\widehat D(u)-D(u)|}{0.05\,\mathrm{m}}\right).
\end{equation}
For position quality, $d_i$ is the distance between the coordinate-wise
medians of the predicted and reference visible 3D surface points in
camera coordinates. We set
$q_i^{\mathrm{pos}}=\exp(-d_i/(0.05\,\mathrm{m}))$.
Both qualities use Eq.~\eqref{eq:supp_complete_quality}.

Object F1 uses the same instance matches:
\begin{equation}
    F_1 = \frac{2N_{\mathrm{TP}}}
    {2N_{\mathrm{TP}}+N_{\mathrm{FP}}+N_{\mathrm{FN}}},
\end{equation}
where $N_{\mathrm{TP}}$ counts matched instances and $N_{\mathrm{FN}}$
counts unmatched reference objects.

For visible-surface F-score, we back-project rendered mesh depth and
reference depth with their respective camera intrinsics to obtain
point sets $P$ and $G$. Both sets are restricted to pixels with valid
reference depth. The metric includes visible surfaces regardless of
instance matching. At $\tau=0.01\,\mathrm{m}$, we compute
\begin{align}
    \mathrm{Prec}_{\tau}
    &= \frac{|\{p\in P:d(p,G)<\tau\}|}{|P|},\\
    \mathrm{Rec}_{\tau}
    &= \frac{|\{g\in G:d(g,P)<\tau\}|}{|G|},\\
    F_{\tau}
    &= \frac{2\,\mathrm{Prec}_{\tau}\mathrm{Rec}_{\tau}}
    {\mathrm{Prec}_{\tau}+\mathrm{Rec}_{\tau}},
\end{align}
where $d(x,A)$ is the distance from $x$ to its nearest point in $A$.
An empty prediction or zero precision and recall yields zero F-score.

\paragraph{Physical fidelity.}
A verified dynamic object is stable if its displacement is below
$1\,\mathrm{cm}$ and its rotation change is below $5^\circ$ after
300 steps under gravity at 240\,Hz (1.25\,s). Stable objects receive
$q_i^{\mathrm{stab}}=1$, and all other objects receive zero.
For a verified dynamic object with annotated supports whose
endpoints are all reconstructed, we compute penetration quality
from its initial penetration depth $p_i$:
\begin{equation}
    q_i^{\mathrm{pen}}
    = \exp\!\left(-\frac{p_i}{0.001\,\mathrm{m}}\right).
\end{equation}
Objects that fail these verification requirements receive zero.
Both scores use Eq.~\eqref{eq:supp_complete_quality}.

\paragraph{Scene-level fidelity.}
Support recall is the fraction of annotated object--support
relations recovered with both endpoints present and the supporting
parent correctly identified. Surface occupancy IoU compares the
voxels occupied by the delivered surfaces, $V_P$, and the reference
surfaces, $V_G$, on a fixed reference grid:
\begin{equation}
    \mathrm{IoU}_{\mathrm{surface}}
    = \frac{|V_P\cap V_G|}{|V_P\cup V_G|}.
\end{equation}

\paragraph{Aggregation.}
We weight scenes equally within each metric's applicable capture set.
Depth quality, position quality, and visible-surface F-score use
HOPE-Video and ARKitScenes captures with reference geometry. Surface
occupancy IoU uses the ARKitScenes references. Failed reconstructions
receive zero for Object F1 and applicable completeness-aware geometry
and physics scores. Visual scores are averaged over available
reference-view renders.

\paragraph{Results.}
The full method leads the reported world-quality comparison across
all four metric groups. Relative to PolaRiS, it improves PSNR from
21.5 to 23.8\,dB and visible-surface F-score from 0.037 to 0.108.
The two ablations also show differences in physical fidelity:
stability is 0.10 with the fixed pipeline and 0.24 without adaptive
refinement, compared with 0.63 for the full method. Support recall
likewise rises from 0.47 and 0.23 in the two ablations to 0.58 in
the full method. The reported gains thus extend from reference-view
appearance to visible geometry, dynamic stability, and support
relations.

\subsection{Physics Calibration}

\paragraph{Tasks and baselines.}
We evaluate parameter estimation from single RGB videos using seven
tasks. EV-RealPhys~\cite{kandukuri2024physics} provides planar sliding
for kinetic friction $\mu_k$. GAUGE~\cite{wang2026gauge} provides
inclined sliding and turntable motion for $\mu_k$, incline slip onset
for static friction $\mu_s$, stick-slip motion for $\mu_k/\mu_s$,
and bouncing for restitution. Pendulum length is evaluated on GAUGE
and IRIS~\cite{khanbayov2026iris}. We use measured geometry or
motion-capture-derived references, depending on the task.

The baselines include PhysX defaults, LLM handbook values,
NIRPI~\cite{hofherr2023nirpi}, Garcia et al.~\cite{garcia2025learning},
2D tracking with analytic dynamics, and direct VLM estimates from
Qwen3-VL-8B and Claude Opus 5.5.

\paragraph{Parameter error.}
For trial $j$ of task $t$, let $\widehat\theta_{t,j}$ and
$\theta^*_{t,j}$ be the estimate and reference. For each method,
$\mathcal V_t$ contains valid numerical point estimates with
references, and $\mathcal T$ contains its scored tasks. We report
the mean of the per-task median absolute relative errors:
\begin{equation}
    E_{\mathrm{param}}
    = \frac{100}{|\mathcal T|}
      \sum_{t\in\mathcal T}
      \operatorname*{median}_{j\in\mathcal V_t}
      \frac{|\widehat\theta_{t,j}-\theta^*_{t,j}|}
           {|\theta^*_{t,j}|}.
\end{equation}
Task coverage and the number of valid trials can vary by method.

\paragraph{Prediction error.}
For a held-out trial, we use the median parameter estimate from other
trials of the same object and surface. Fixed-parameter baselines use
their assigned values. Analytic dynamics with shared extracted
observations predict four outcomes: the endpoints of planar and
inclined slides, the duration of the flight after a bounce, and
the equivalent small-angle pendulum period. For slides, we estimate
the initial position and velocity from the first 40\% of the free-motion
segment and normalize endpoint error by the remaining observed travel.
For bouncing, restitution scales the preceding flight duration.
For pendula, $T_0=2\pi\sqrt{L/g}$, with the held-out trial's fitted length
defining the reference period. We average per-task median relative
errors over the tasks with valid predictions.

\paragraph{Bound violation.}
We report the percentage of evaluated outputs that contradict static
friction constraints implied by observed sticking or sliding.
For example, sticking on an incline of angle $\alpha$ implies
$\mu_s\geq\tan\alpha$. Baseline point estimates and our conservative
one-sided bounds are checked according to their respective output
forms. Dashes indicate unavailable or inapplicable results.

\paragraph{Results.}
Our method covers all seven parameter tasks and achieves 4.2\%
parameter error and 2.6\% prediction error. The 2D tracking baseline
reports 24.4\% and 13.1\%, respectively, over its supported tasks.
This comparison favors our method on the reported aggregates,
whose task coverage varies across methods. The 0.0\% bound violation
rate indicates that our reported friction bounds are consistent
with the observed motion constraints. On the evaluated subset,
estimates also vary little under mirroring or reductions in
resolution or frame rate. This diagnostic measures sensitivity
through the median relative change between valid estimates before
and after perturbation.

\end{document}